\documentclass{article}
\usepackage{spconf,amsmath,graphicx,hyperref}

\usepackage{array}
\usepackage{multirow}
\usepackage{makecell} 
\usepackage{float}
\usepackage{booktabs}

\newcolumntype{M}[1]{>{\centering\arraybackslash}m{#1}}
\newcolumntype{L}[1]{>{\raggedright\arraybackslash}m{#1}}

\title{Structure-to-Image: Zero-Shot Depth Estimation in Colonoscopy via High-Fidelity Sim-to-Real Adaptation}
\name{Juan Yang$^{1}$ \qquad Yuyan Zhang$^{2}$ \qquad Han Jia$^{1}$ \qquad Bing Hu$^{2,\ast}$ \qquad Wanzhong Song$^{1,\ast}$\thanks{$^{\ast}$Corresponding authors. \newline \indent E-mail: songwz@scu.edu.cn (W. Song), hubing@wchscu.edu.cn (B. Hu).}}

\address{$^{1}$College of Computer Science, Sichuan University, Chengdu, China \\
         $^{2}$Department of Gastroenterology and Hepatology, West China Hospital, Sichuan University, Chengdu, China}

\begin{document}
\ninept
\maketitle
\begin{abstract}

Monocular depth estimation (MDE) for colonoscopy is hampered by the domain gap between simulated and real-world images. Existing image-to-image translation methods, which use depth as a posterior constraint, often produce structural distortions and specular highlights by failing to balance realism with structure consistency. To address this, we propose a “Structure-to-Image” paradigm that transforms the depth map from a passive constraint into an active generative foundation. We are the first to introduce phase congruency to colonoscopic domain adaptation and design a cross-level structure constraint to co-optimize geometric structures and fine-grained details like vascular textures. In zero-shot evaluations conducted on a publicly available phantom dataset, the MDE model that was fine-tuned on our generated data achieved a maximum reduction of 44.18\% in RMSE compared to competing methods. Our code is available at https://github.com/YyangJJuan/PC-S2I.git.

\end{abstract}
\begin{keywords}
Domain Adaptation, Depth Estimation, \\Colonoscopy, Structure-to-Image, Phase Congruency
\end{keywords}
\section{Introduction}
\label{sec:intro}

Colorectal cancers (CRC) are the third most reported incidence rate and the second highest mortality rate worldwide\cite{araghi2019global}. Colonoscopy is the screening gold standard. Its effectiveness is operator-dependent, leading to a polyp miss-rate of approximately 20\%\cite{vemulapalli2022most}.Intra-procedural 3D map, enabled by dense depth estimation\cite{ma2019real,schops2019surfelmeshing}, can address this limitation by ensuring complete inspection\cite{wan2021polyp,wu2024toder}.

Monocular depth estimation (MDE) in colonoscopy is hindered by a domain gap\cite{kim2024density}, as models is usually trained on synthetic data due to the absence of real-world ground truth\cite{chen2021beyond,he2024monolot,solano2025multi}. The low fidelity of synthetic data in texture and lighting causes poor generalization, making Sim-to-Real adaptation crucial for clinical accuracy\cite{mahmood2018deep}.

\begin{figure}[t!]
    \centering
    \includegraphics[width=0.85\columnwidth]{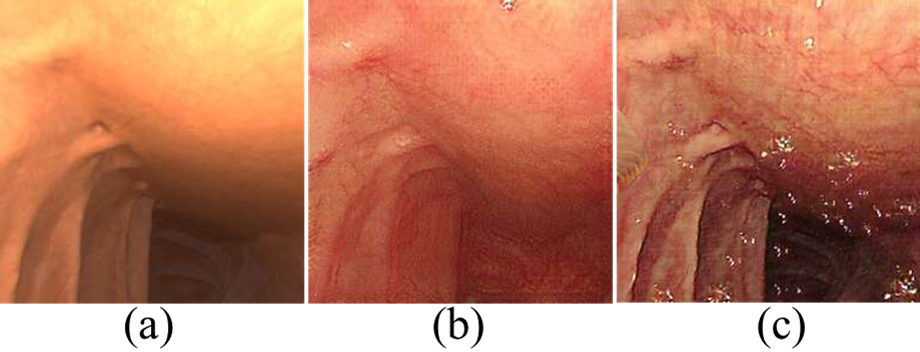}
    \caption{Results of existing CycleGAN-based methods for Sim-to-Real colonoscopy show structural distortions and specular artifacts.}
    \label{fig:Fig1example}
\end{figure}

CycleGAN-based Sim-to-Real is one of the primary approaches to this challenge\cite{zhu2017unpaired,jeong2021depth}, following two main pathways: translating synthetic images into realistic ones, or translating depth-encoded virtual colonoscopy grayscale images into realistic images. However, both pathways are prone to structural distortions. To preserve structure, Wang et al.\cite{wang2024structure} used mutual information between the source depth map and the generated realistic image; Tomasini et al.\cite{tomasini2024sim2real} used a pre-trained MDE model to enforce depth consistency with the source map; and Mathew et al.\cite{mathew2020augmenting} proposed XDCycleGAN to strengthen geometric constraints via an extended cycle consistency loss and directional discriminators. However, existing image-to-image methods still produce local distortions and specular artifacts ( as shown in Fig. \ref{fig:Fig1example}), limiting the performance of downstream MDE models.

Sim-to-Real colonoscopy image generation focuses on infusing synthetic images with clinical realism while preserving accurate depth labels\cite{kim2023effective}. Real colonoscopy images, however, contain not only macro-structures\cite{bustos2013colon} that are represented in depth maps (e.g., lumen, folds, polyps), but also fine-grained micro-structures\cite{judge2025texture} (e.g., sub-mucosal vascular patterns) that are absent from them. Existing image translation methods often face the challenge in balancing the synthesis of realistic micro-structures with the accurate preservation of underlying macro-structural geometry. 

We argue that within the context of Sim-to-Real colonoscopy image generation, depth information should function not merely as an auxiliary constraint, but as the foundational structural basis. We propose a “Structure-to-Image” strategy that trains a CycleGAN to generate realistic appearances from cross-level inputs. This approach shifts the paradigm from “preserving depth during image translation” to “generating realistic appearance from a structural foundation”. We introduce cross-level structure constraint that enforces spatial consistency with the input depth and micro-structural similarity to real images. The main contributions are:

(1) “Structure-to-Image” generation paradigm is proposed that elevates structure from a passive constraint to the generative foundation, enhancing the geometric accuracy and image realism.

(2) A novel cross-level structure constraint is designed to optimize the spatial geometry and micro-structures during generation.

(3) Zero-shot depth estimation evaluations on phantom dataset verified that the proposed method achieved up to 44.18\% reduction in RMSE compared to competing methods.

\begin{figure*}[t!]
    \centering
    \includegraphics[width=0.99\textwidth]{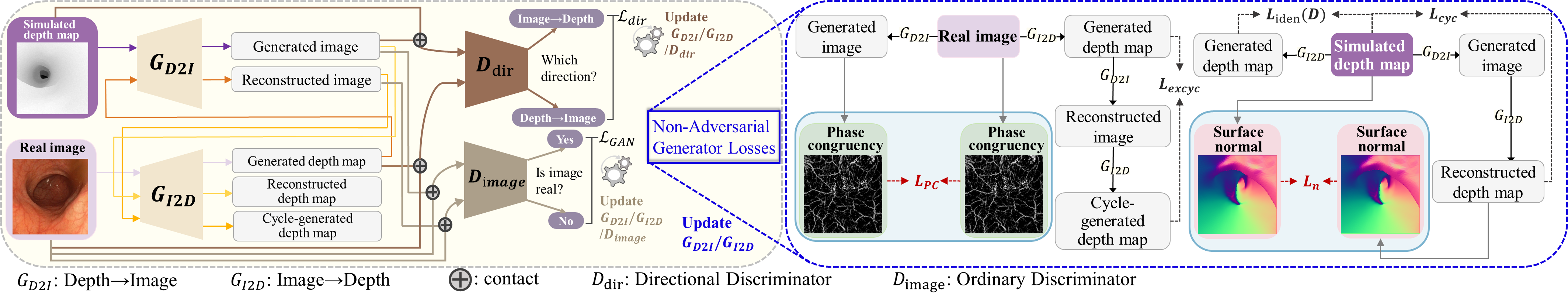}
    \caption{Structure-to-Image training pipeline with a cross-level structure constraint for geometric and micro-structural alignment}
    \label{fig:ModelTraining}
\end{figure*}

\section{Method}
\label{sec:method}
\begin{figure}[t!]
    \centering
    \includegraphics[width=0.82\columnwidth]{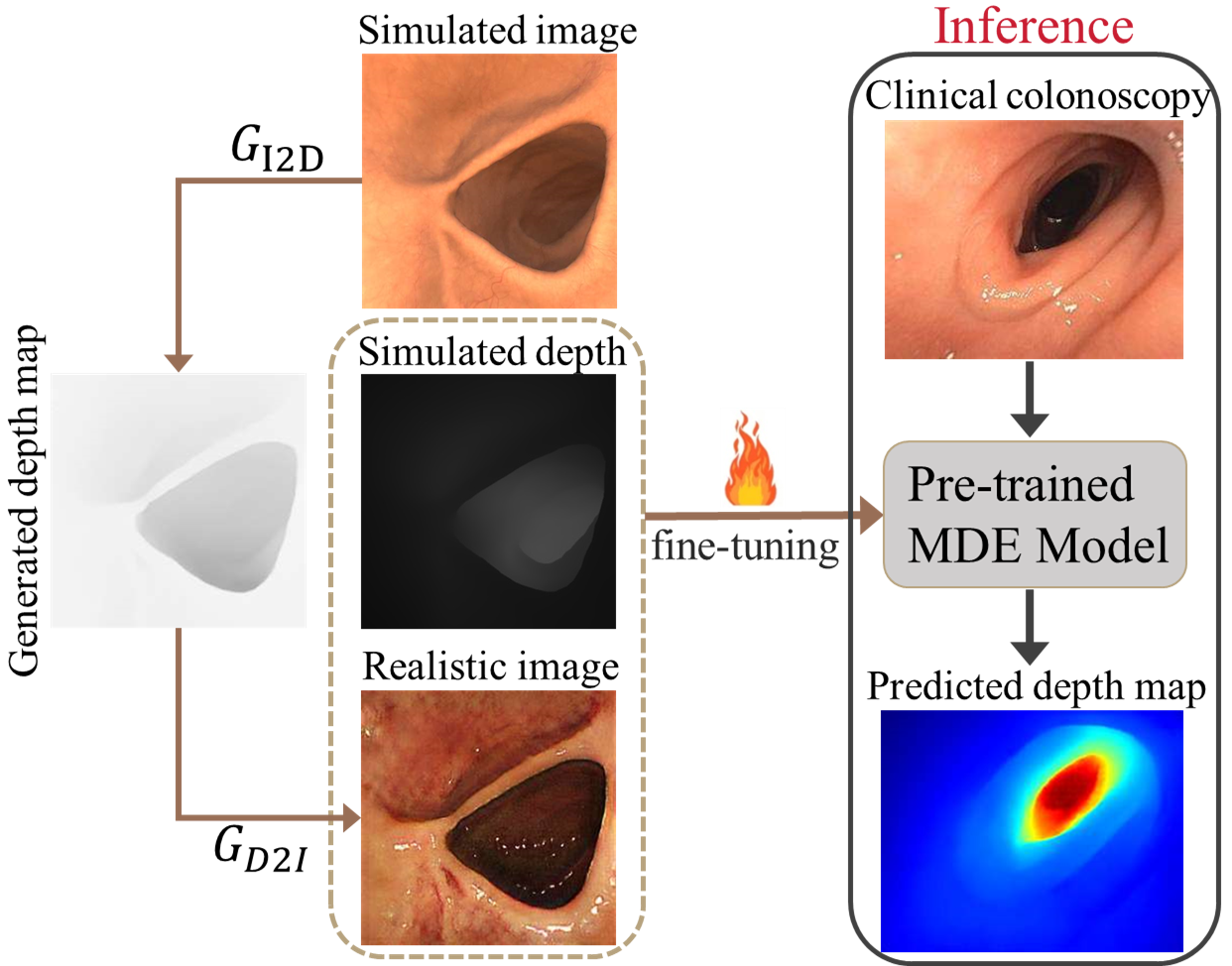}
    \caption{Inference process of the proposed method.}
    \label{fig:Modelinference}
\end{figure}

Our framework (Fig.\ref{fig:ModelTraining}) trains a CycleGAN\cite{zhu2017unpaired} on unpaired real images and synthetic depth maps, guided by a cross-level structure constraint to align geometric and micro-structural features. The trained generator (Fig.\ref{fig:Modelinference}) generates realistic images from generated depth maps, forming paired image-simulated depth that are used to fine-tune the downstream MDE model and enhance zero-shot depth accuracy on real-world colonoscopy.

\subsection{Structure-to-Image}

Unlike image-to-image methods that struggle with the trade-off between realism and structure consistency, our structure-to-Image strategy treats structure as a generative prior, not a posterior constraint. This simplifies the generator's task from “simultaneously inferring structure and appearance" to “generating a matching appearance for a given structure," which fundamentally reduces learning uncertainty and enhances generation stability.

\begin{figure}[t!]
    \centering
    \includegraphics[width=0.97\columnwidth]{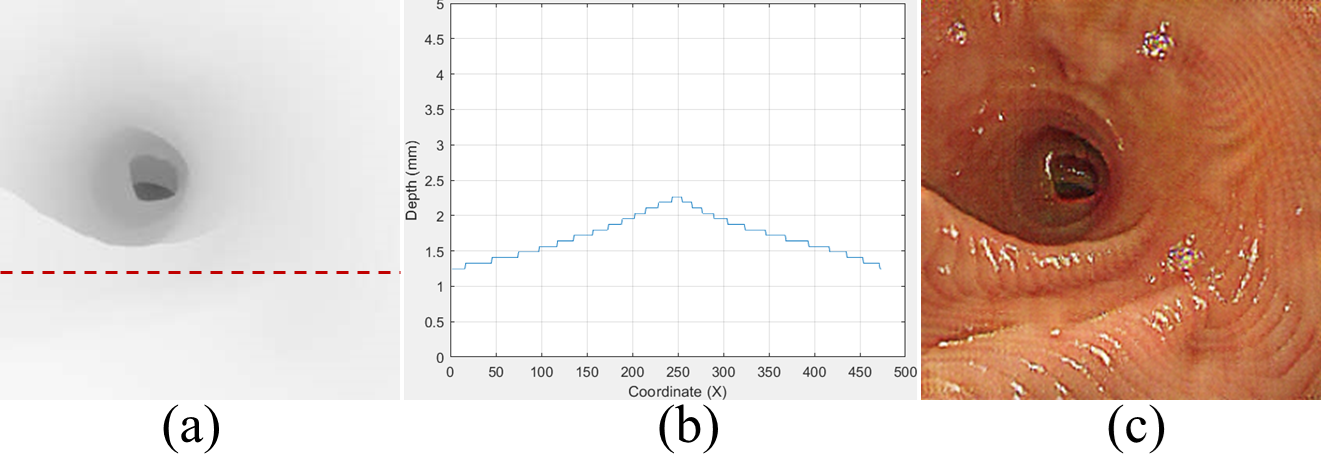}
    \caption{Stair-step depth map causes contour-like artifacts in the generated image. (a) Depth map of the SimCol dataset, (b) depth profile along the red dashed line of (a), (c) generated realistic image from (a) with depth-to-image translation.}
    \label{fig:Stair-step}
\end{figure}

Our Structure-to-Image translation task relies on accurate depth maps as input. However, depth maps of public simulated and phantom colonoscopy datasets contain stair-like depth values, which introduce contour-like distortions in the generated images (Fig. \ref{fig:Stair-step}). 

Inspired by the work of XDCycleGAN\cite{mathew2020augmenting}, we construct a unified generative framework (Fig.\ref{fig:ModelTraining}) which can simultaneously trains two branches: an image-to-depth branch that generates accurate depth maps, and a depth-to-image branch that generates realistic images. This unified framework can achieves depth estimation performance comparable to specialized models like NormDepth\cite{wang2023surface}.

To satisfy XDCycleGAN's input requirement for inverse depth, we transform the 16-bit positive depth maps into inverse depth maps:
\begin{equation}
    D_{-} = 1 - D_{+}/65535,
\end{equation}
where $D_{+}$ is the positive depth, and $D_{-}$ is the inverse depth. 

Although the framework's image-to-depth module can estimate depth, its goal is not to achieve state-of-the-art accuracy, as GANs typically achieve inferior performance in MDE when compared to supervised learning-based models\cite{rau2019implicit,cheng2021depth,paruchuri2024leveraging}. Instead, this module is designed to generate depth maps that are structurally sufficient for our Structure-to-Image task, which in turn serves the core goal of narrowing the sim-to-real domain gap.

\subsection{Cross-level Structure Constraint}

Generating realistic colonoscopy images requires the simultaneous preservation of accurate geometric structures\cite{bustos2013colon} (e.g., lumen, folds, polyps) and fine-grained micro-structures\cite{judge2025texture} (e.g., vascular textures). This is challenging due to inherent appearance ambiguity (for instance, a dark area could be a shadow or pathological tissue), which can misguide the model and thus necessitates a contrast-insensitive, unified structural descriptor.

\subsubsection{Phase Congruency Loss}

\begin{figure}[t!]
    \centering
    \includegraphics[width=0.97\columnwidth]{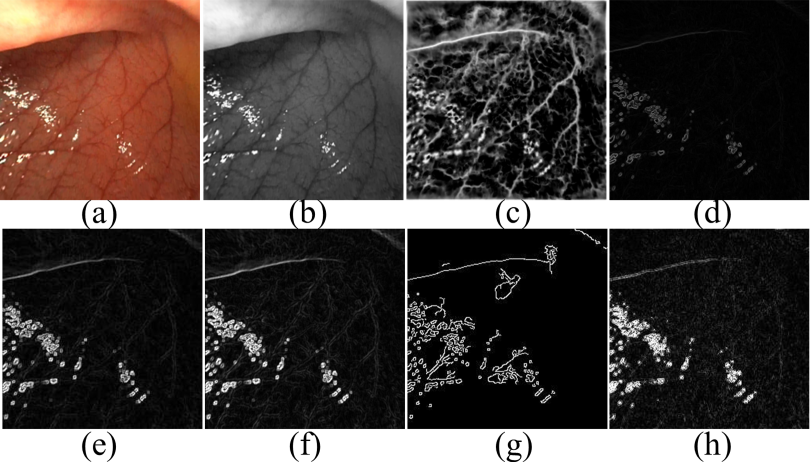}
    \caption{Structures extracted with various methods. (a) real colonoscopy image; (b) Y-Channel of (a); (c) phase consistency map of (a); (d)-(h) edges extracted using Roberts, Prewitt, Sobel, Canny, and Laplacian operators, respectively. Phase congruency map shows more and accurate structures. }
    \label{fig:PCmap}
\end{figure}


Existing works have shown that the important structural information in an image appears at the locations where the phase of its Fourier components reaches maximum consistency\cite{marr1980theory,morrone1986mach,morrone1987feature,henriksson2009representation}. Phase Congruency (PC) can simultaneously locate geometric and micro-structures in the frequency domain\cite{kovesi1999image}. As shown in Fig.\ref{fig:PCmap}, compared with traditional edge detection operators, PC can more robustly extract micro-structural details (such as blood vessels), while preserving the macroscopic geometric contours.

Inspired by this, we propose a cross-level structure constraint, phase congruency loss. 
We convolve the input image with a set of multi-scale and multi-orientation Log-Gabor filters\cite{field1987relations} to obtain a series of filter responses. Phase congruency is computed\cite{zhang2011fsim} as
\begin{equation}
    PC(x) = \frac{\sum_{j} E_{\theta_j}(x)}{\epsilon + \sum_{n}\sum_{j} A_{n,\theta_j}(I)},
\end{equation}
where $x$ is a point of the image, $E_{\theta_j}(x)$ is the local energy of the responses along orientation $\theta_j$, $A_{n,\theta_j}$ is the local amplitude of the responses on scale $n$ and orientation $\theta_j$, $\epsilon$ is a small positive constant.

For image $I_{real}$ and image $I_{gen}$ ($ I_{gen}= G_{D2I}(I_{real}))$, the phase congruency loss is
\begin{equation}
    \mathcal{L}_{PC} = 1 - \frac{\sum_x S_{PC}(x)S_{G}(x)PC_{m}(x)}{\sum_x PC_{m}(x)},
\end{equation}
where $PC_{m}(x) = \max(PC_{gen}(x), PC_{real}(x))$, and 
\begin{align}
    S_{PC}(x) &= \frac{2PC_{gen}(x) \cdot PC_{real}(x)+T_{1}}{PC_{gen}^{2}(x)+PC_{real}^{2}(x)+T_{1}} \\
    S_{G}(x) &= \frac{2G_{gen}(x) \cdot G_{real}(x)+T_{2}}{G_{gen}^{2}(x)+G_{real}^{2}(x)+T_{2}}
\end{align}
$T_1=0.85$, $T_2=160$\cite{zhang2011fsim} . $G_{rel}$ and $G_{gen}$ are the gradient magnitude of $I_{real}$ and $I_{gen}$. Gradient magnitude is used to enhance the geometric constraint.

\subsubsection{Normal Consistent Loss}
To further align fine geometric structures, the normal consistent loss is used to constraint the normal of simulated depth map $Depth_{sim}$ and the depth map $Depth_{rec}$ as
\begin{equation}
    \mathcal{L}_{n} = \frac{1}{N} \sum \left( 1 - \frac{\vec{n}_{sim} \cdot \vec{n}_{rec}}{\| \vec{n}_{sim} \|_2 \cdot \| \vec{n}_{rec} \|_2} \right),
\end{equation}
where $\vec{n}_{sim}$ and $\vec{n}_{rec}$ are the normal of $Depth_{sim}$ and $Depth_{rec}$, respectively. $Depth_{rec} = G_{I2D}(G_{D2I}(Depth_{sim}))$.

The total loss function is
\begin{equation}
    \begin{aligned}
        \mathcal{L} = & \alpha\mathcal{L}_{GAN} + \beta\mathcal{L}_{cyc} + \beta\mathcal{L}_{excyc}  + \mathcal{L}_{dir} + \gamma\mathcal{L}_{iden}(D) + \\ &\gamma\mathcal{L}_{PC} + \lambda\mathcal{L}_{n},
    \end{aligned}
\end{equation}
where $\mathcal{L}_{GAN}$, $\mathcal{L}_{cyc}$, and $\mathcal{L}_{iden(D)}$ are the same as CycleGAN. $\mathcal{L}_{excyc}$ and $\mathcal{L}_{dir}$ are the same as XDCycleGAN. In this study, $\alpha=0.5$, $\beta=10$, $\gamma=5$, and $\lambda=2.4$.

\begin{figure}[t!]
    \centering
    \includegraphics[width=0.99\columnwidth]{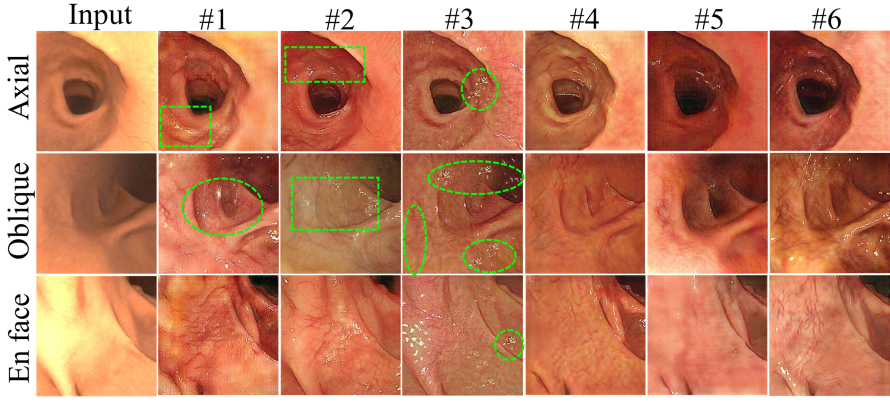}
    \caption{Comparison of generated realistic images. Boxes denote structural errors; circles denote specular artifacts. Methods: \#1-CycleGAN, \#2-Struct-Preserve, \#3-Sim2Real, \#4-Ours-w/o Normal, \#5-Ours-w/o Phase, \#6-Ours.}
    \label{fig:generateImage}
\end{figure}

\begin{figure*}[t!] 
    \centering
    \includegraphics[width=0.99\textwidth]{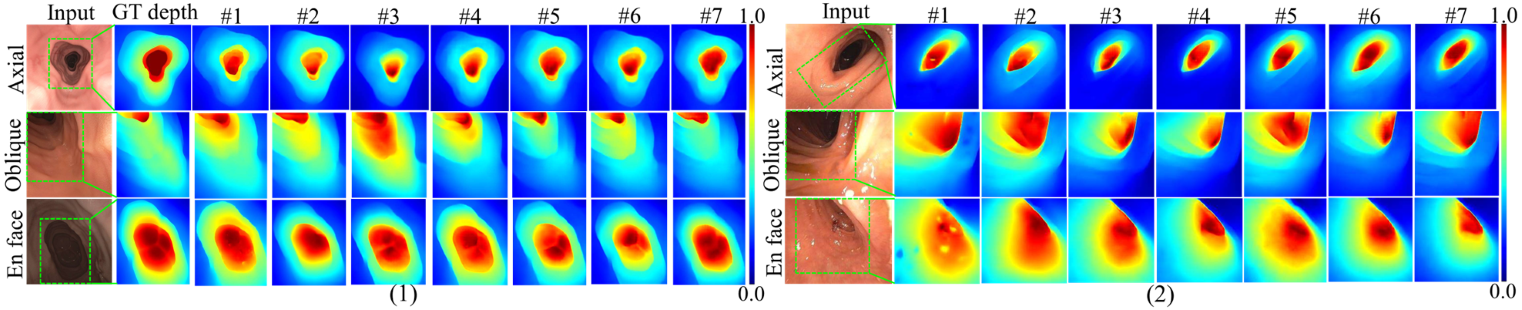} 
    \caption{Qualitative comparison of zero-shot depth estimation on C3VD (phantom) and Colon10K (real) datasets. An pre-trained MDE model was fine-tuned on datasets from each method listed. Green boxes highlight key prediction differences. Methods: \#1-Baseline, \#2-XDCycleGAN, \#3-Struct-Preserve, \#4-Sim2Real, \#5-Ours-w/o Normal, \#6-Ours-w/o Phase, \#7-Ours.}
    \label{fig:DAV2-image}
\end{figure*}

\section{Experiments}
\label{sec:experiments}

\subsection{Implementation Details}
We use four colonoscopy datasets, including one simulated dataset (SimCol\cite{rau2023bimodal}), one phantom dataset (C3VD\cite{bobrow2023colonoscopy}), and two real-world datasets (Colon10K\cite{ma2021colon10k} and Colon-Ours). 

Building upon the XDCycleGAN framework and its original hyperparameters\cite{mathew2020augmenting}, our model is trained for 200 epochs. We set the weight for the loss $\mathcal{L}_{n}$ to $\lambda=2.4$ to balance realism with structural consistency. In later training stages, the loss $\mathcal{L}_{PC}$, with a weight of $\gamma=5$, was applied only after epoch 160 to ensure stability of the image-to-depth training phase.
All the code was implemented in Python 3.12 and CUDA 12.4 using PyTorch 2.5.1. Our training is conducted on one RTX 4090 GPU using a batch size of 6 and a learning rate of $2\times10^{-4}$. All comparison methods used their original hyper-parameters.

Metrics of single-frame depth accuracy in Sect. ~\ref{sec:Comparison of Image-to-Depth Accuracy} and ~\ref{sec:Evaluation of Downstream Depth Estimation} include root mean square error (RMSE), mean absolute error (MAE), and squared relative error (SqRel).

\subsection{Comparison of Realistic Image Generation}
\label{sec:Comparison of Realistic Image Generation}

We compare our method with three image-to-image colonoscopy methods, including XDCycleGAN\cite{mathew2020augmenting}, Struct-Preserve\cite{wang2024structure}, and Sim2Real\cite{tomasini2024sim2real}. Two wo variants (Ours-w/o Phase, Ours-w/o Normal) of our method are also included. 
All models were trained on a dataset combining Colon-Ours (2,083~frames) and SimCol depth maps (3,800~frames). After training, models are used to generate $256 \times 256$ realistic images from SimCol RGB images (37,833~frames). Evaluation metrics includes peak signal-to-noise ratio (PSNR), structural similarity index measure (SSIM), and inception score (IS). We did not use Fr\'{e}chet Inception Distance because our stylistically diverse clinical data lacks a coherent reference distribution.

Table \ref{tab:image_quality_assessment} lists the quantitative results, confirming our method's superior performance on both structural (PSNR/SSIM) and quality (IS) metrics. Fig. \ref{fig:generateImage} shows the qualitative results, which reveal the structure-realism trade-off inherent in competing methods. In contrast, our approach successfully generates realistic textures and physically plausible highlights while preserving the geometric structure.

\begin{table}[htbp]
    \centering
    \caption{Quantitative image generation results. \textbf{Bold}: Best result; \underline{Underline}: Second-best.}
    \label{tab:image_quality_assessment}
    \begin{tabular}{l c c c} 
        \toprule 
        \textbf{Methods} & \textbf{PSNR$\uparrow$} & \textbf{SSIM$\uparrow$} & \textbf{IS$\uparrow$} \\
        \midrule 
        XDCycleGAN\cite{mathew2020augmenting} & 17.70 & 0.61 & \underline{3.38$\pm$0.15} \\
        Struct-Preserve\cite{wang2024structure} & 16.68 & 0.50 & 3.20$\pm$0.11 \\
        Sim2Real\cite{tomasini2024sim2real} & 17.76 & 0.66 & 2.92$\pm$0.08 \\
        Ours-w/o Phase & \underline{19.89} & \underline{0.72} & 3.10$\pm$0.09 \\
        Ours-w/o Normal & 19.84 & 0.66 & 3.21$\pm$0.09 \\
        Ours & \textbf{20.65} & \textbf{0.74} & \textbf{3.47$\pm$0.14} \\
        \bottomrule 
    \end{tabular}
\end{table}

\subsection{Comparison of Image-to-Depth Accuracy}
\label{sec:Comparison of Image-to-Depth Accuracy}
Both XDCycleGAN\cite{mathew2020augmenting} and ours can directly generate depth maps from RGB images. To check the performance of this image-to-depth scheme. We conducted zero-shot depth generation on the phantom dataset (C3VD) according to the partition\cite{rodriguez2023lightdepth}. Both XDCycleGAN and our model are trained on two separate datasets, the first comprising 10,000 SimCol image-depth pairs, and the second a mix of 2,083 Colon-Ours RGB images and 3,800 SimCol depth maps. All models are test on C3VD dataset. Another method, NormDepth\cite{wang2023surface}, that reported its zero-shot depth accuracy on C3VD metrics is included in this comparison. Table \ref{tab:c3vd_MDE_zero_shot} lists the comparison results. 

The performance gap between XDCycleGAN and XDCycleGAN-C in Table \ref{tab:c3vd_MDE_zero_shot} highlights its sensitivity to training data, with poor generalization causing structural anomalies that hinder clinical application. In contrast, our method maintains stability across training conditions, outperforms XDCycleGAN, and achieves performance comparable to NormDepth, demonstrating its capacity to provide robust depth estimation and reliable structural priors.

\begin{table}[htbp]
    \centering
    \caption{Zero-shot depth generation errors on C3VD dataset of image-to-depth generative models. \textbf{Bold}: Best result; \underline{Underline}: Second-best. Models trained on Colon-Ours are marked with “-C".}
    \label{tab:c3vd_MDE_zero_shot}
    \begin{tabular}{lccc}
        \toprule 
       \textbf{Model} & \makecell{\textbf{RMSE$\downarrow$} \\ \textbf{(mm)}} & \makecell{\textbf{MAE$\downarrow$} \\ \textbf{(mm)}} & \makecell{\textbf{SqRel$\downarrow$} \\ \textbf{(mm)}} \\
        \midrule 
        XDCycleGAN\cite{mathew2020augmenting} & 7.74 & 5.51 & 1.40 \\
        XDCycleGAN-C\cite{mathew2020augmenting} & 15.00 & 8.89 & 8.66 \\
        NormDepth\cite{wang2023surface} & \textbf{7.41} & - & 2.88 \\
        Ours-w/o Phase & 7.60 & 5.46 & \underline{1.38} \\
        Ours-w/o Normal & 7.60 & \underline{5.32} & 1.44 \\
        Ours-C & 9.25 & 6.04 & 2.98 \\
        Ours & \underline{7.53} & \textbf{5.28} & \textbf{1.32} \\
        \bottomrule 
    \end{tabular}
\end{table}

\subsection{Evaluation of Downstream Depth Estimation}
\label{sec:Evaluation of Downstream Depth Estimation}
To further verify the effectiveness of the proposed method. We select the pre-trained DepthAnythingV2-small model\cite{yang2024depth} as the foundation model of the downstream MDE task. Six datasets are built for fine-tuning the foundation model, with each dataset corresponding to a separately fine-tuned model. These six datasets are built using methods listed in the Table \ref{tab:image_quality_assessment}. Baseline is the model fine-tuned in a supervised manner on SimCol dataset. All seven models, including Baseline, perform zero-shot depth inference on phantom dataset (C3VD, 10,015~frames) and real dataset (Colon10K, 3,000~frames).

Table \ref{tab:depth_anything_comparison} lists the zero-shot depth estimation errors on C3VD. Our method surpasses other six methods; for instance, it reduces RMSE by 25.95\% compared to Baseline, and by 32.60\% and 44.18\% compared to Struct-Preserve and Sim2Real, respectively. This demonstrates that realistic images generated with our method can enhance the accuracy of SOTA depth models for colonoscopy.

\begin{table}[htbp]
    \centering
    \caption{Zero-shot depth estimation errors on C3VD dataset of fine-tuned MDE models. \textbf{Bold}: Best result; \underline{Underline}: Second-best.}
    \label{tab:depth_anything_comparison}
    \begin{tabular}{lccc}
        \toprule 
        \textbf{Model} & \makecell{\textbf{RMSE$\downarrow$} \\ \textbf{(mm)}} & \makecell{\textbf{MAE$\downarrow$} \\ \textbf{(mm)}} & \makecell{\textbf{SqRel$\downarrow$} \\ \textbf{(mm)}} \\
        \midrule 
        Baseline & 6.28 & 4.27 & 0.97 \\
        XDCycleGAN\cite{mathew2020augmenting} & 7.27 & 4.81 & 1.19 \\
        Struct-Preserve\cite{wang2024structure} & 6.90 & 4.70 & 1.13 \\
        Sim2Real\cite{tomasini2024sim2real} & 8.33 & 5.58 & 1.84 \\
        Ours-w/o Phase & \underline{4.94} & \underline{3.34} & \underline{0.67} \\
        Ours-w/o Normal & 5.15 & 3.42 & 0.71 \\
        Ours & \textbf{4.65} & \textbf{3.15} & \textbf{0.62} \\
        \bottomrule 
    \end{tabular}
\end{table}

Fig. \ref{fig:DAV2-image} displays depth maps predicted with the seven models on C3VD (phantom) and Colon10K (real) datasets. On Colon10K (b), Baseline fails by misinterpreting specular highlights as structure, confirming the limits of simple synthetic data. In contrast, the model fine-tuned with our data accurately capturing both overall structures like Haustral folds and fine local details, such as the en-face intestinal wall in (a) and the oblique-view folds in (b).

\subsection{Ablation study}
As shown in Table \ref{tab:image_quality_assessment}-Table \ref{tab:depth_anything_comparison}, ablation studies validate the effectiveness of the cross-level structure constraint. The performance of our two variants, Ours-w/o Phase (lacking the phase consistency constraint) and Ours-w/o Normal (lacking the normal consistency constraint), confirms that both components independently enhance the model's accuracy on the core domain adaptation task.


\section{Conclusion}
\label{sec:conclusion}

Our “Structure-to-Image" strategy, using a cross-level constraint to co-optimize realism and structural consistency, enables the downstream MDE model to achieve a maximum 44.18\% RMSE reduction over competitors in zero-shot inference. Relying on predicted depth avoids synthetic artifacts but introduces prediction bias as a limitation, which future work will address by building a smooth synthetic dataset and researching controllable vascular texture generation.

\section{Acknowledgment}
\label{sec:acknowledgment}
This work was supported by the National Natural Science Foundation of China (Grant No. 62576227).







\let\oldbibliography\thebibliography
\renewcommand{\thebibliography}[1]{%
  \oldbibliography{#1}%
  \setlength{\itemsep}{1pt} 
  \setlength{\parskip}{0.5pt}  
  \setlength{\parsep}{0.5pt}   
}

\bibliographystyle{IEEEbib}
\bibliography{refs}

\end{document}